\pdfoutput=1

\documentclass[11pt]{article}

\usepackage{acl}

\usepackage{times}
\usepackage{url}
\usepackage{latexsym}

\usepackage[T1]{fontenc}

\usepackage[utf8]{inputenc}

\usepackage{microtype}

\usepackage{amsmath,amssymb,amsfonts,latexsym,dsfont,xspace}
\usepackage{physics,nicefrac,mathtools}
\usepackage{multirow}
\usepackage{verbatim}
\usepackage{xcolor}
\usepackage{graphicx}
\usepackage{subcaption}
\usepackage{array}
\usepackage{booktabs}
\usepackage{colortbl}
\usepackage{pgfplotstable}
\usepackage[normalem]{ulem} %
\usepackage{graphicx}
\usepackage{float}

\usepackage{todonotes}
\setuptodonotes{fancyline, color=red!30, size=\scriptsize}

\usepackage{diagbox}
\usepackage{wrapfig}

\newif\iffinal
\finaltrue

\newif\ifediting
\editingfalse

\ifediting
\newcommand{\fnb}[1]{{\color{red}\footnote{\textcolor{red}{B: #1}}}}
\newcommand{\fnw}[1]{{\color{blue}\footnote{\textcolor{blue}{W: #1}}}}

\newcommand\rmw{\bgroup\markoverwith
{\textcolor{blue}{\rule[0.5ex]{2pt}{1pt}}}\ULon}
\newcommand\rmb{\bgroup\markoverwith
{\textcolor{red}{\rule[0.5ex]{2pt}{1pt}}}\ULon}
\else  %
\newcommand{\fnb}[1]{}
\newcommand{\fnw}[1]{}

\newcommand{\rmb}[1]{}
\newcommand{\rmw}[1]{}
\fi

\newcommand{\flores}[0]{\textsc{Flores}}

\DeclareMathOperator*{\optop}{top-T}
\DeclareMathOperator*{\argmax}{arg\,max} %
\newcommand{\hypset}{\ensuremath{\bar{\mathcal{H}}(x)}}

\newcommand{\hypsetT}{\ensuremath{\bar{\mathcal{H}}_T(x)}}

\newcommand{\one}{\textit{i)}\xspace}
\newcommand{\two}{\textit{ii)}\xspace}
\newcommand{\three}{\textit{iii)}\xspace}
\newcommand{\four}{\textit{iv)}\xspace}

\newcommand{\mbrnn}{MBR$_{\text{N-by-N}}$\xspace}
\newcommand{\mbrns}{MBR$_{\text{N-by-S}}$\xspace}
\newcommand{\mbrcf}{MBR$_{\text{C2F}}$\xspace}
\newcommand{\ie}{\textit{i.e.}\xspace}
\newcommand{\eg}{\textit{e.g.}\xspace}

\newcommand{\wrt}{w.r.t.\xspace}

\newcommand{\topparam}{\ensuremath{T}\xspace}

\newcommand{\nbest}{$k$-best\xspace}
\newcommand{\beamsize}{$k$\xspace}

\newcommand{\ymap}{\ensuremath{y^{\text{MAP}}}} %
\newcommand{\ymbr}{\ensuremath{y^{\text{MBR}}}} %
\newcommand{\ycoling}{\ensuremath{y^{\text{N-by-N}}}} %
\newcommand{\ycf}{\ensuremath{y^{\text{C2F}}}}

\title{Sampling-Based Approximations to Minimum Bayes Risk Decoding for Neural Machine Translation}

\author{Bryan Eikema \\
  University of Amsterdam \\
  {\tt b.eikema@uva.nl} \\\And
  Wilker Aziz \\
  University of Amsterdam \\
  {\tt w.aziz@uva.nl} \\}

\begin{document}
\maketitle
\begin{abstract}
In NMT we search for the mode of the model distribution to form predictions.  
The mode and other high-probability translations found by beam search have been shown to often be inadequate in a number of ways.
This prevents improving translation quality through better search, as these idiosyncratic translations end up selected by the decoding algorithm, a problem known as the beam search curse.
Recently, an approximation to minimum Bayes risk (MBR) decoding has been proposed as an alternative decision rule that would likely not suffer from the same problems. We analyse this approximation and establish that it has no equivalent to the beam search curse.  
We then design approximations that decouple the cost of exploration from the cost of robust estimation of expected utility. 
This allows for much larger hypothesis spaces, which we show to be beneficial. 
We also show that mode-seeking strategies can aid in constructing compact sets of promising hypotheses and that MBR is effective in identifying good translations in them.
We conduct experiments on three language pairs varying in amounts of resources available: English into and from German, Romanian, and Nepali.\footnote{Code is available at \url{github.com/roxot/mbr-nmt}.}
\end{abstract}

\begin{figure}[t]
    \centering
    \includegraphics[width=1.0\columnwidth]{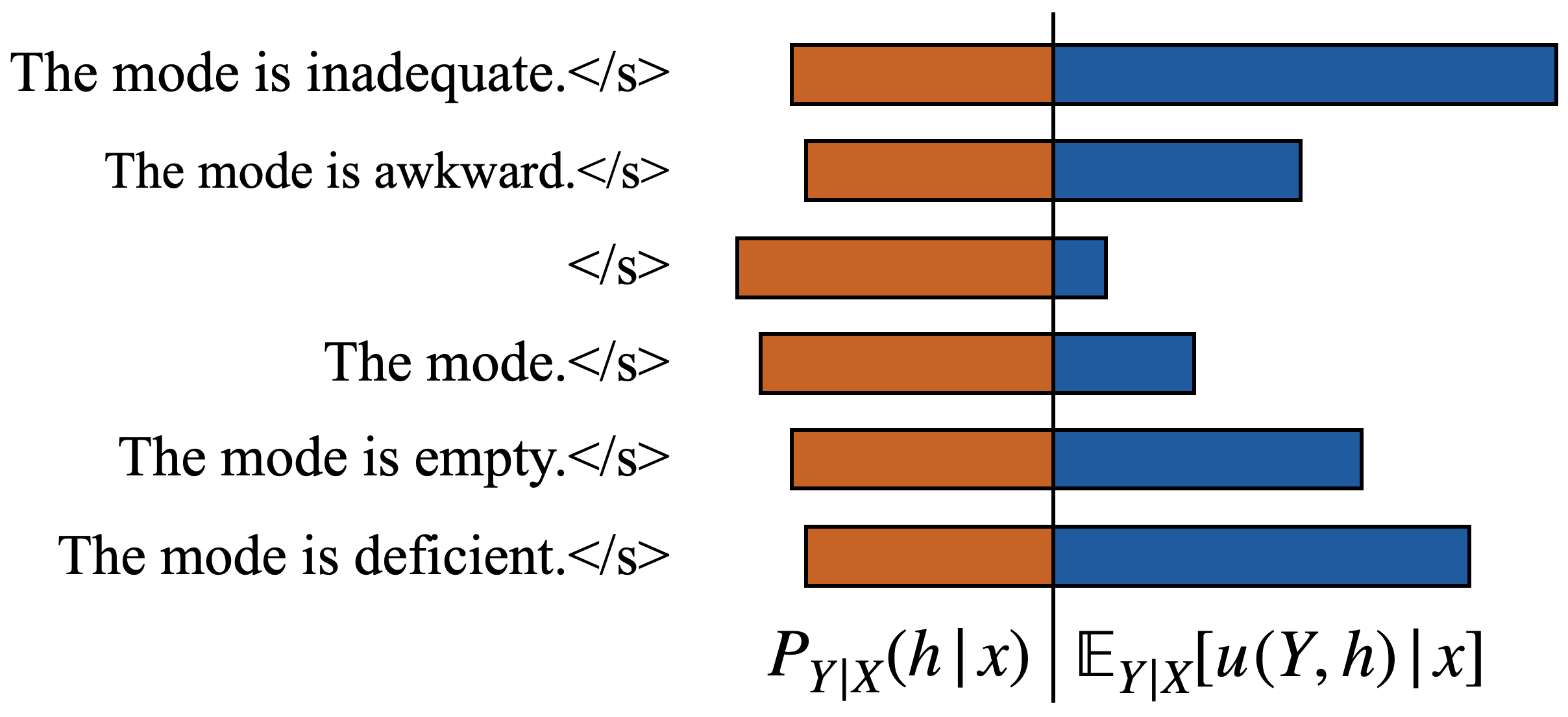}
    \caption{NMT spreads probability roughly uniformly over a large set of promising hypotheses (left). %
    MBR (right) assigns hypotheses an expected utility, %
    revealing clear preferences against those that are too idiosyncratic. %
    } 
    \label{fig:MBR}
\end{figure}

\section{Introduction}

NMT systems \citep{sutskever-etal-2014-sequence,bahdanau-etal-2015-nmt} are trained to predict a conditional probability distribution over translation candidates of any given source sentence. After training, choosing a translation for a given input requires a decision rule: a criterion to elect a `preferred' translation. MAP decoding, the most common decision rule in NMT, seeks the most probable translation under the model (\ie, the mode of the  distribution).

MAP decoding and its approximations such as beam search \citep{beamsearch-graves} have been under scrutiny. 
\citet{stahlberg-byrne-2019-nmt} show that the true mode %
is oftentimes inadequately short or empty. 
Better approximate search is known to hurt quality \citep{koehn-knowles-2017-six,murray-chiang-2018-correcting,kumar-sarawagi-2019-calibration}, a problem known as the \emph{beam search curse}.
The success of beam search depends on search biases introduced by hyperparameters such as beam size %
and length normalisation, which are tuned not to correlate with the objective of MAP decoding, but rather to strike a compromise between mode-seeking search and properties of reasonable translations. %
Despite its success, a number of problems have been observed: length bias \citep{cho-etal-2014-properties,sountsov-sarawagi-2016-length}, word frequency bias  \citep{ott-etal-2018-analyzing}, susceptibility to copy noise \citep{khayrallah-koehn-2018-impact,ott-etal-2018-analyzing}, and hallucination under domain shift \citep{lee2019hallucinations,muller-etal-2020-domain,wang-sennrich-2020-exposure}.

\citet{eikema-aziz-2020-map} argue that the inadequacy of the mode in NMT is a reasonable consequence of the translation space being combinatorial and unbounded. 
They show that, while distributions predicted by NMT do reproduce various statistics of observed data, they tend to spread probability mass almost uniformly over a large space of translation candidates. %
This makes their precise ranking in terms of probability mass a fragile criterion for prediction.
While some of these candidates are possibly inadequate (\eg, the empty sequence), most of them are similar to one another and
exhibit appreciable structural similarity to reference translations.  
To make better use of the statistics predicted by NMT models, \citet{eikema-aziz-2020-map} recommend MBR decoding \citep{kumar-byrne-2004-minimum}, a decision rule that seeks the translation candidate which maximises an external notion of utility (\eg, an MT evaluation metric) in expectation under the model distribution.
While MBR decoding promises robustness to %
idiosyncratic translations, it remains intractable, much like MAP decoding. 
\citet{eikema-aziz-2020-map} propose an approximation based on Monte Carlo (MC) sampling, which although tractable in principle, requires a prohibitive number of assessments of the utility function.

In this work, we first analyse the procedure by \citet{eikema-aziz-2020-map} and establish that it does not suffer from a counterpart to the beam search curse.
That is, better search does not hurt translation quality.
Their approximation is, however, computationally expensive, requiring a number of assessments of the utility function that is quadratic in sample size.
We propose algorithms that scale linearly, allowing us to explore large hypothesis spaces, and considerably improve upon existing approximations to MBR with less computation.
Finally, we find that mode-seeking strategies such nucleus sampling and beam search can still aid MBR decoding by constructing compact sets of high expected utility hypotheses, relying on MBR to filter idiosyncratic translations that may be present.

\section{NMT and Decision Rules}
\label{sec:background}

NMT employs neural networks (NNs)  to predict a conditional probability distribution $Y|\theta, x$ over translation candidates of any given source sentence $x$. %
 The sample space $\mathcal Y$ is the set of all sequences of known target-language symbols (\eg, sub-word units). 
NMT factorises the distribution as a chain of random draws from Categorical distributions
\begin{equation}\label{eq:story}
    Y_j|\theta, x, y_{<j} \sim \mathrm{Cat}(f(x, y_{<j}; \theta)) 
\end{equation}
parameterised in context.
The prefix translation $y_{<j}$ starts empty and grows one symbol at a time until a special end-of-sequence symbol is drawn. At each step $j$, $f$ maps from varying inputs $(x, y_{<j})$ to a probability distribution over the vocabulary. 
Common choices for $f$ include recurrent networks \citep{sutskever-etal-2014-sequence,bahdanau-etal-2015-nmt} %
and Transformers \citep{transformer}.
Given a dataset of observed translation pairs, the NN parameters $\theta$ are estimated to attain a local optimum of the regularised log-likelihood function. %

After training, and for a given input, 
choosing a translation requires a \textit{decision rule} to map from a distribution over translation candidates to a single `preferred' translation. %
The most common decision rule in NMT is MAP decoding, which outputs the mode of the conditional distribution. %
Despite the widespread intuition that MAP decoding is an obvious choice, maximum likelihood estimation (MLE) is oblivious to our desire to form predictions. %

\subsection{MAP Decoding}

Maximum-a-posteriori (MAP) decoding outputs the most probable translation under the model:\footnote{The name is a historical accident, NMT models directly parameterise the conditional distribution without the need for a prior, and, thus, without posterior inference.}
\begin{equation}\label{eq:MAP}
    \ymap = \argmax_{h \in \mathcal Y} ~ \log p_{Y|X}(h|x,\theta) ~.
\end{equation}
As this is intractable, beam search \citep{beamsearch-graves,sutskever-etal-2014-sequence} is used. Beam search is a pruned version of breadth-first search which maintains an active set of \beamsize partial translations. %
For large beam size \beamsize, translation quality degrades  \citep{koehn-knowles-2017-six} and the exact $\ymap$ is often the empty sequence \citep{stahlberg-byrne-2019-nmt}. Therefore, in practice, the beam size is kept small and the objective in Equation (\ref{eq:MAP}) is regularised to up-rank longer hypotheses \citep{Wu2016GooglesNM,murray-chiang-2018-correcting}.

\subsection{MBR Decoding}
\label{sec:background:mbr}

Minimum Bayes risk (MBR) decoding stems from the principle of maximisation of expected utility  \citep{BergerSDT}. 
A utility function $u(y, h)$ measures the benefit in choosing $h \in \mathcal Y$ when $y \in \mathcal Y$ is the ideal decision.
When forming predictions, we lack knowledge about ideal translations and must decide under uncertainty. 
MBR lets the model %
fill in `ideal decisions' probabilistically as we search through the space of candidates for the one which is assigned highest utility \emph{in expectation}:
\begin{equation}\label{eq:MBR}
\ymbr = \argmax_{h \in \mathcal Y} ~ \underbrace{\mathbb E[u(Y, h) \mid \theta, x]}_{\eqqcolon \mu_u(h; x, \theta)} ~.
\end{equation}
MBR has a long history in parsing \citep{goodman-1996-parsing,simaan-2003-maximizing}, speech recognition \citep{stolcke1997explicit,goelbyrnembr}, and MT \citep{kumar-byrne-2002-minimum,kumar-byrne-2004-minimum}. %

In MT, $u$ can be a sentence-level evaluation metric (\eg, METEOR~\citep{meteor} or 
Sentence BLEU~\citep{chen-cherry-2014-systematic}).
Intuitively, %
whereas the MAP prediction is the translation to which the model assigns highest probability, no matter how idiosyncratic,  the MBR prediction is the translation that is closest (under the chosen~$u$) to all other probable translations.%
 See Figure~\ref{fig:MBR} for an illustration of this concept.
Seeking support for a prediction not only in terms of probability but also in terms of utility makes MBR decoding robust to situations where inadequate translations are  assigned high probability, as it often happens with the empty string \citep{stahlberg-byrne-2019-nmt}, when the training data are noisy \citep{ott-etal-2018-analyzing},  too small \citep{eikema-aziz-2020-map} or distant from the test domain \citep{muller2021understanding}.

It is a well-known result that for the `exact match' utility, $u(y,h) \coloneqq \mathbf{1}_{\{y\}}(h)$, the expected utility of $h$ is $p_{Y|X}(h|x,\theta)$, hence MBR and MAP decoding have the same optimum under this choice \citep{kumar-byrne-2002-minimum}. This view justifies MAP decoding as an instance of MBR, where decisions are optimised with respect to a strict notion of translational equivalence. In machine translation evaluation, exact match is a questionable choice of utility function. It, for example, is unable to capture paraphrases or any other form of semantic equivalence. 

Like in MAP decoding, exhaustive enumeration of all hypotheses is impossible, we must resort to a finite subset $\hypset$ of candidates.
Unlike MAP decoding, the objective function $\mu_u(h; x, \theta)$ \emph{cannot} be evaluated exactly.
Most approximations to MBR decoding, from  \citet{kumar-byrne-2004-minimum} to recent instances \citep{stahlberg-etal-2017-neural,shu-nakayama-2017-later,blain2017exploring}, use \nbest lists from beam search for $\hypset$ and to form a biased estimate of expected utility. 
\citet{eikema-aziz-2020-map} use unbiased samples from the model for both approximations: \one they follow the generative story in Equation (\ref{eq:story}) to obtain $N$ independent samples $y^{(n)}$ , a procedure known as ancestral sampling \citep{MCbook}; %
then, \two for a hypothesis $h$, they compute an MC estimate of $\mu_u(h;x,\theta)$:
\vspace{-0.5em}
\begin{equation}\label{eq:MC}
    \hat\mu_u(h; x, N) \overset{\text{MC}}{\coloneqq} \frac{1}{N} \sum_{n=1}^N u(y^{(n)}, h) ~,
\end{equation}
which is unbiased for any sample size $N$.
\citet{eikema-aziz-2020-map} use the same $N$ samples as candidates and approximate Equation (\ref{eq:MBR}) by 
\begin{equation}
\label{eq:sampling-based-mbr}
    \ycoling \coloneqq \argmax_{h \in \{y^{(1)}, \ldots, y^{(N)}\}} ~\hat\mu_u(h;x,N) ~ . %
\end{equation}
We note that the candidates do not need to be obtained using ancestral sampling, and  investigate alternative strategies in Section~\ref{sec:experiments:alternative-hypothesis-spaces}. It is important, however, to use ancestral samples to obtain an unbiased estimate of expected utility as we show in Section~\ref{sec:experiments:estimation-bias}. We call this class of MBR algorithms using unbiased MC estimation instances of \emph{sampling-based MBR decoding}.

\section{Coarse-to-Fine MBR Decoding}
\label{sec:method}

\begin{figure*}[t]
    \centering
    \scriptsize
    \setlength{\tabcolsep}{1ex}
    \begin{tabular}{ l p{0.95\textwidth}}
        \midrule
        src & Convercent erhielt \$10 Millionen bei der Finanzierung im Februar von Firmen wie Sapphire Ventures und Tola Capital, womit das gesamte Kapital auf \$47 Millionen angehoben wurde.  \\
        ref & Convercent raised \$10 million in funding in February from firms such as Sapphire Ventures and Tola Capital, bringing its total capital raised to \$47 million. \\ 
        \midrule
    \end{tabular}
    \includegraphics[width=\textwidth]{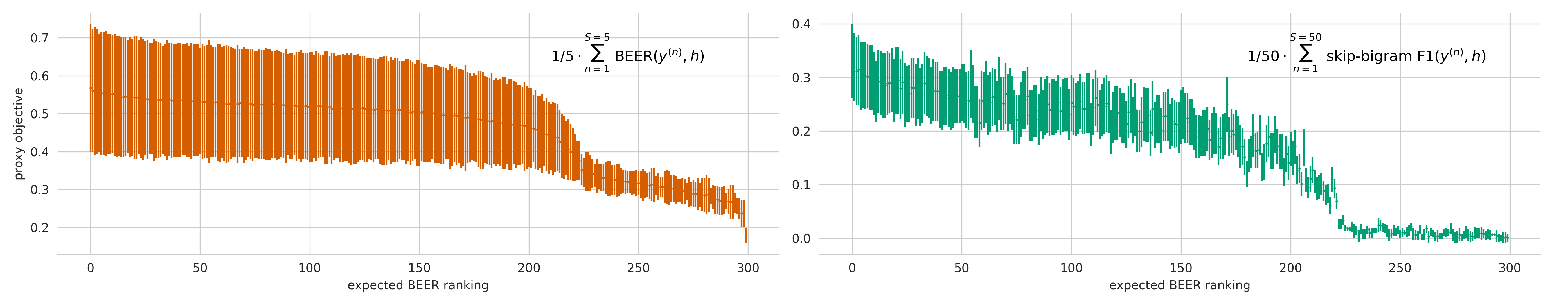}
    \vspace{-1em}
    \caption{Motivation for coarse-to-fine MBR. We sort 300 candidates sampled from the model along the x-axis from best to worst according to a robust MC estimate (using 1,000 samples) of expected BEER under the model. 
    Left: feasible MC estimates (5 samples) of each candidate's expected BEER.
    Right: robust and inexpensive MC estimates (100 samples) of expected utility \wrt a simpler metric (skip-bigram F1).  As estimates are stochastic, we perform 100 repetitions and plot mean $\pm$ two  deviations. We can see that the robust estimates (right) correlate fairly well with the expensive ranking we intend to approximate (x-axis), despite of the simpler utility. As we can afford more evaluations of the proxy utility, we obtain estimates of reduced variance, which leads to safer pruning.}
    \label{fig:mbr-filter-strategies}
\end{figure*} 

A big disadvantage of \mbrnn is that it requires $N^2$ assessments of the utility function. If $U$ is an upperbound on the time necessary to assess the utility function once, then \mbrnn runs in time $\mathcal O(N^2 \times U)$.
For a complex utility function, this can grow expensive even for a modest hypothesis space.
As NMT distributions have been shown to be high entropy~\citep{ott-etal-2018-analyzing,eikema-aziz-2020-map}, the quadratic cost prevents us from sufficiently exploring the space of translations. Therefore, we investigate and propose more flexible algorithms. 

An important property of sampling-based MBR decoding is that MC estimation of expected utility, Equation~(\ref{eq:MC}), and approximation of the hypothesis space in Equation~(\ref{eq:sampling-based-mbr}) really are two independent approximations. Tying the two is no more than a design choice that must be reconsidered. 
We start by obtaining $N$ translation candidates from the model, which will form the hypothesis space \hypset. Then, we use any  number $S < N$ of ancestral samples for approximating expected utility in Equation~(\ref{eq:MC}).\footnote{In practice, for efficiency we will use a fixed set of $S$ samples to estimate expected utility for each candidate.}
We call this version \mbrns, which takes time $\mathcal O(N \times S \times U)$. %
Compared to \mbrnn, this variant is able to scale to much larger hypothesis spaces $\hypset$. %
In practice, however, robust MC estimation for the utility of interest may still require $S$ that is too large for the $N$ we are interested in.

An idea that we explore in this work is to make use of a proxy utility that correlates with the target utility but is cheaper to compute. Even when those do not correlate perfectly, we can make use of the proxy utility to filter the hypothesis space to a manageable size \topparam on which we can perform robust MC estimation of expected utility.  We coin this approach coarse-to-fine MBR decoding (or \mbrcf), which filters the hypothesis space to a manageable size in the coarse step, and performs robust MC estimation of expected utility in the fine step: 
\begin{subequations}\label{eq:mbr-c2f}
\begin{align}
    \ycf &\coloneqq \argmax_{h \in \hypsetT} ~\hat\mu_{u_{\text{target}}}(h;x,L) \label{eq:mbr-c2f-fine} \\
    \hypsetT &\coloneqq \optop_{h \in \hypset} ~ \hat \mu_{u_{\text{proxy}}}(h; x, S) ~. \label{eq:mbr-c2f-coarse}
\end{align}    
\end{subequations}
Upper-bounding the complexity of the proxy utility by $U_\text{proxy}$, the target utility by $U_\text{target}$, using $S$ samples for MC estimation in the coarse step (\ref{eq:mbr-c2f-coarse}) and $L$ in the fine step (\ref{eq:mbr-c2f-fine}), the complexity of this algorithm is $\mathcal O(N \times S \times U_{\text{proxy}} + T \times L \times U_{\text{target}})$. 
\mbrcf  decouples robust MC estimation (large $L$) from exploration (large $N$) and the cost of exploration from the cost of the target utility.

As illustrated in Figure~\ref{fig:mbr-filter-strategies}, we can find proxy utilities that correlate reasonably well with our target utility and are able to give us a rough---but useful---ordering of the hypothesis space. Rather than using a proxy utility, we could use the target utility itself in the coarse-step provided we pick a small $S$. This, however, most likely leads to too high variability in the ranking, as shown in Figure~\ref{fig:mbr-filter-strategies} (left). %

\section{Data, Systems and Utilities}

We perform experiments on three language pairs with varying amount of resources for training: English into and from German, Romanian and Nepali.
For German-English (de-en) we use all available WMT'18~\citep{bojar-etal-2018-findings} news data except for Paracrawl, resulting in 5.9 million sentence pairs. We train a Transformer base model~\citep{transformer} until convergence and average the last 10 epoch checkpoints to obtain our final model. We test our models on \texttt{newstest2018}.  
For Romanian-English (ro-en) we use all available WMT'16~\citep{bojar-etal-2016-findings} news data amounting to 565k sentence pairs. We train a Transformer base model until convergence and pick the best epoch checkpoint according to the validation loss. We test our models on \texttt{newstest2016}.
Finally, for Nepali-English (ne-en) we use the data setup by \citet{guzman-etal-2019-flores}. We apply the pre-processing step of removing duplicates as in \citet{eikema-aziz-2020-map}. This results in 235k sentence pairs. We test our models on the \flores{} test set, which is of a widely different domain than the training data. We mimick the training setup and models used in \citet{guzman-etal-2019-flores}.
In all models we disable label smoothing, as this has been found to negatively impact model fit, which would compromise the performance of MBR~\citep{eikema-aziz-2020-map}.

For computational efficiency, we opt for non-neural evaluation metrics for use as utility function in MBR. BEER~\citep{stanojevic-simaan-2014-fitting} is a non-neural trained metric that has shown good correlation with human judgements in previous WMT metrics shared tasks~\citep{machacek-bojar-2014-results,stanojevic-etal-2015-results,bojar-etal-2016-results}. 
In experiments shown in Table~\ref{tab:app:utilities} in Appendix~\ref{app:additional-results} we found that using BEER as utility function performed well at pushing translation performance higher across a range of automatic evaluation metrics. We therefore use BEER as the utility of choice in our experiments and as a consequence will consistently report corpus-level BEER scores of MBR translations as well. We also report SacreBLEU~\citep{bleu,sacrebleu} scores where relevant to be able to detect overfitting to the utility and for comparison with other works.

\section{Experiments}

\subsection{Estimation of Expected Utility}
\label{sec:experiments:estimation-bias}

\begin{figure}[t]
    \centering
    \includegraphics[width=0.48\textwidth]{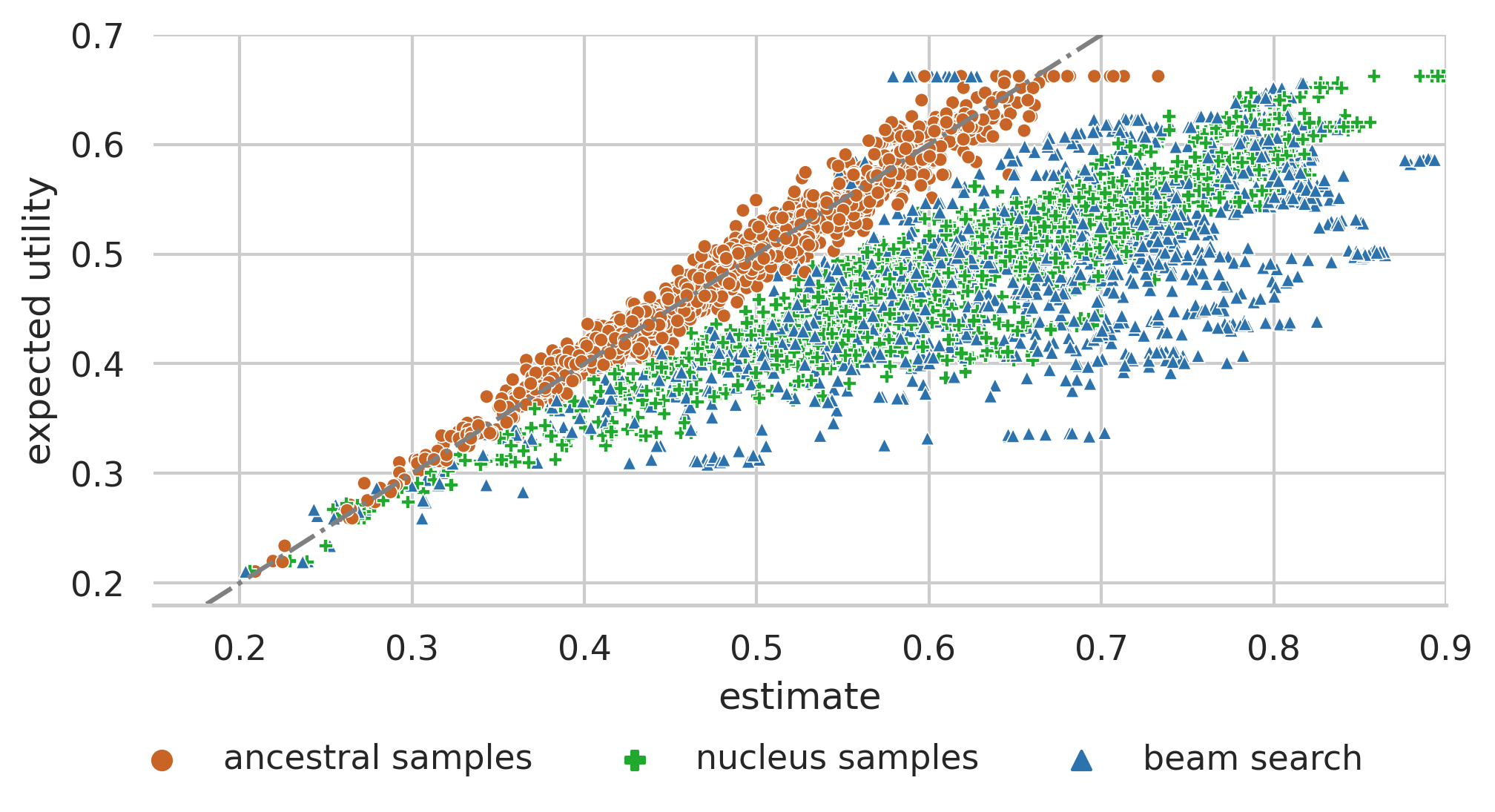}
    \caption{
    Estimates of expected utility for various hypotheses. We plot practical estimates of expected utility (x-axis) using either ancestral, nucleus or `beam' samples against an accurate MC estimate using 1,000 ancestral samples. The gray line depicts a perfect estimator.
    }
    \label{fig:exp-utility-estimation-bias}
\end{figure}

We start by motivating the importance of unbiased estimates of expected utility using ancestral samples (\ie sampling-based MBR). In Figure \ref{fig:exp-utility-estimation-bias} we verify the biasedness of alternatives to ancestral sampling for this computation: nucleus sampling \citep{holtzman-etal-2019-the} and `beam sampling' (\ie, using $k$-best outputs from beam search for estimating expected utility; \citet{blain2017exploring}). We can see, rather clearly, that estimates using nucleus samples or beam search bias away from expected utility under the model, while ancestral sampling is unbiased by design and hence should be preferred when approximating the objective function in search. Therefore, in all experiments that follow, we shall use ancestral samples for making unbiased estimates of expected utility, even when different methods are used to construct the hypothesis space.

\subsection{N-by-N MBR}
\label{sec:experiments:n-by-n-mbr}

\begin{figure}[t]
    \centering
    \includegraphics[width=\columnwidth]{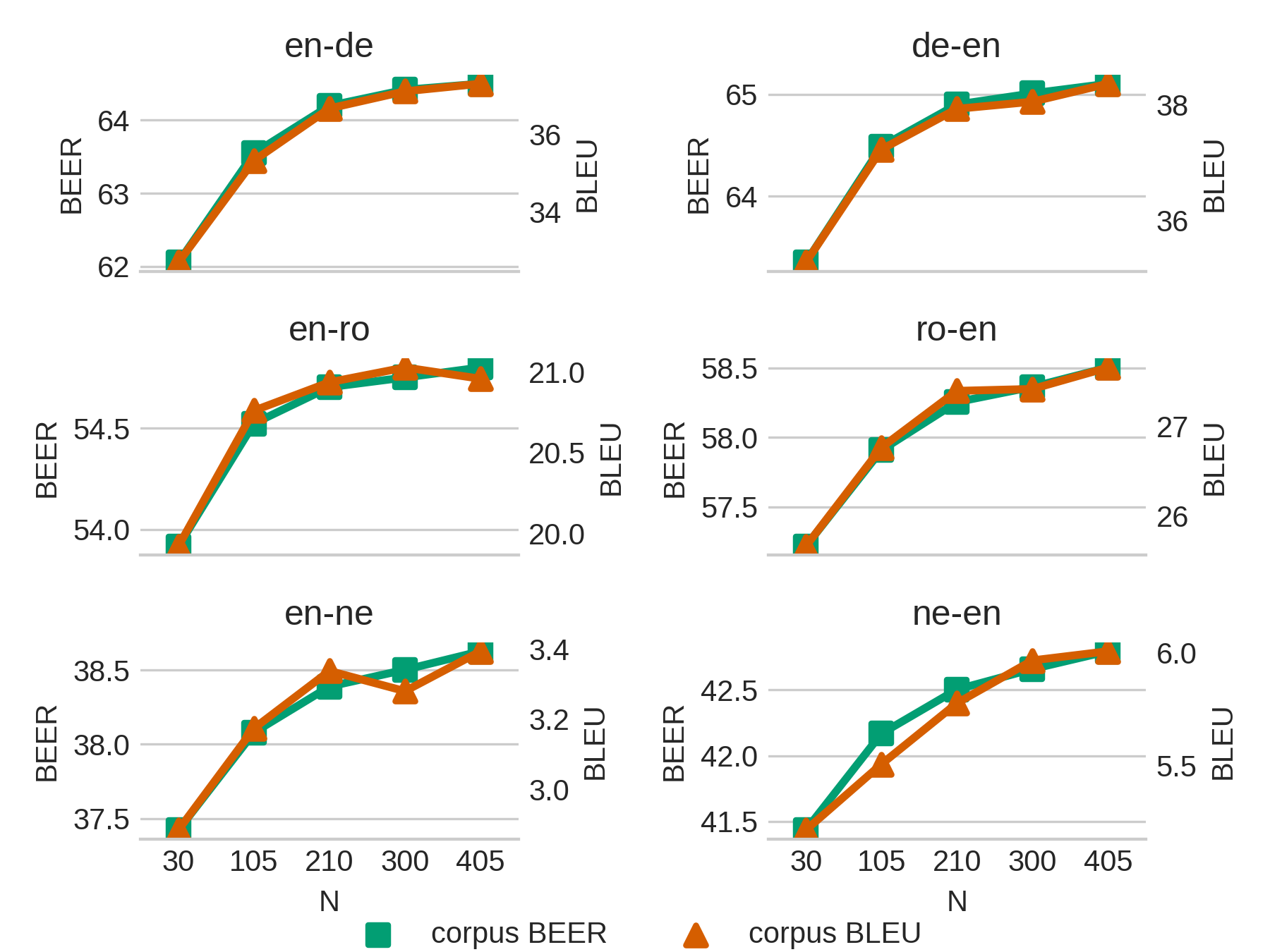}
    \caption{\mbrnn for various sizes of $N$ using BEER as target utility. We report both BEER and BLEU scores.}
    \label{fig:nbyn}
\end{figure}

Now, we look into scaling \mbrnn. %
\citet{eikema-aziz-2020-map} only explored 30 by 30 approximations to the MBR objective. Our aim is to investigate whether MBR decoding is indeed able to scale to better translation performance with more computation. 
In Figure~\ref{fig:nbyn}, we explore $N$ from 30 to 405.\footnote{A batch size of 15 is convenient on our hardware, which is why we work with multiples of 15 in most experiments.} As MBR optimises a specific utility (we use BEER), we report translation quality along both BEER and BLEU to detect overfitting to the metric.

We find that MBR steadily improves across language pairs as $N$ grows larger. BLEU scores improve at a similar rate to that of BEER, showing no signs of overfitting to the utility. This is strong empirical evidence that \emph{sampling-based} MBR  has no equivalent to the beam search curse. 
We see this as an important property of a decoding objective.

\subsection{N-by-S MBR}
\label{sec:experiments:n-by-s-mbr}

\begin{figure}[t]
    \centering
    \includegraphics[width=\columnwidth]{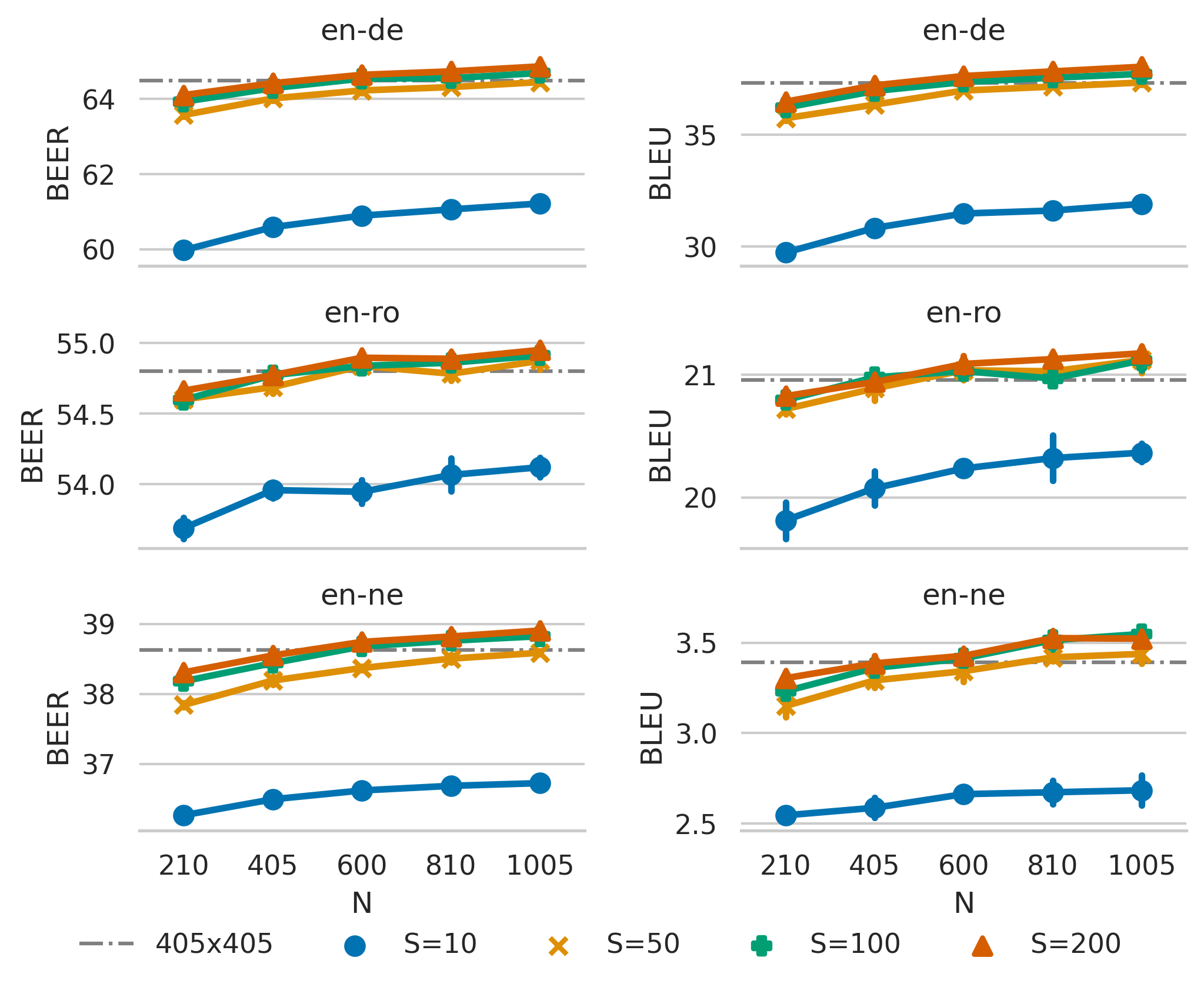}
    \caption{\mbrns: we estimate the expected utility of $N$ hypotheses using $S$ samples. We show average performance over 3 runs with 1 standard deviation. The dashed line shows \mbrnn performance at $N=405$.} %
    \label{fig:mbr-n-by-s}
\end{figure}

\mbrnn couples two approximations, namely, tractable exploration and unbiased estimation of expected utility are based on the same $N$ ancestral samples. %
Our aim is to learn more about the impact of these two approximations, for which 
we look into \mbrns. 
Moreover, with less than $N^2$ assessments of utilities per decoding, we can also investigate larger $\hypset$. We explore $N$ ranging from 210 to 1005, while keeping the number of samples used for approximating expected utility of each hypothesis smaller, with $S$ ranging from 10 to 200. We argue that $S$ does not need to grow at the same pace as $N$, as MC estimates should stabilize after a certain point.\footnote{The standard error of the mean scales with the inverse square root of the sample size.} See our results in Figure~\ref{fig:mbr-n-by-s}. %

We find that growing $N$ beyong $405$ improves translation quality further, even when the estimates of expected utility are less accurate. 
Increasing $S$ also steadily improves translation quality, with diminishing returns in the magnitude of improvement.
On the other hand, smaller values of $S$ lead to notable deterioration of translation quality and we note higher variance in results. 
For all language pairs it is possible to improve upon the best \mbrnn results by considering a larger hypothesis spaces and smaller $S$. %
This experiment shows that the two approximations can be controlled independently and better results are within reach if we explore more. %
On top of that, the best setting of \mbrnn takes 164,025 utility assessments per decoding, \mbrns with $S=100$ brings this number down to 100,500 for the largest $N$ considered, while improving BEER scores on all language pairs.
We note that again increasing either $N$ or $S$ generally improves translation quality in our experiments. This further strengthens our previous finding that sampling-based MBR does not seem to have an equivalent of the beam search curse.

\subsection{Choice of Hypothesis Space}
\label{sec:experiments:alternative-hypothesis-spaces}
\begin{figure*}
    \centering
    \includegraphics[width=\textwidth]{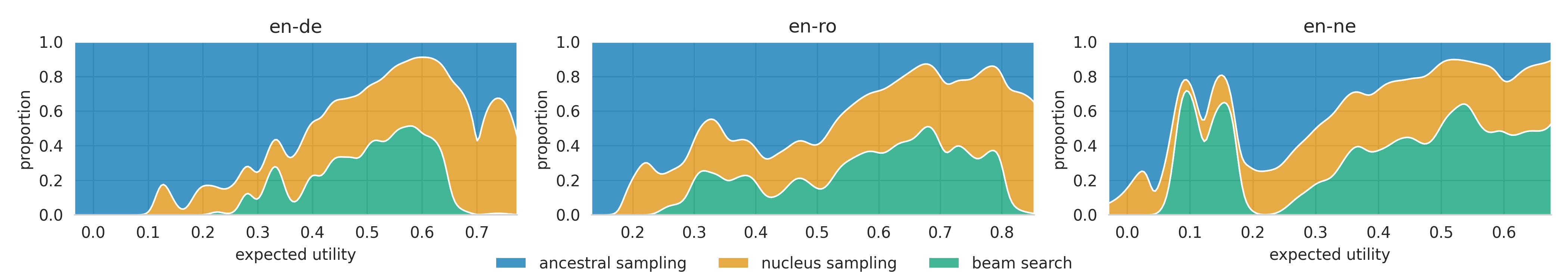}
    \caption{Proportion plots of expected utility for 3 strategies for constructing $\hypset$, using 100 translation candidates per strategy. We estimate expected utility using 1,000 samples. Results are aggregated over 100 source sentences.}
    \label{fig:alternative-hypothesis-spaces}
\end{figure*}

While our focus thus far has been on reducing the number of target utility calls, allowing the exploration of larger $\hypset$, one should also take sampling time in consideration. For example, we found that in \mbrnn with $N=100$, sampling time made up about $60\%$ of the total translation time for our setup. Therefore, it is computationally attractive to construct compact $\hypset$ with promising translation candidates. Ideally, for better search in MBR, we enumerate a set of high expected utility hypotheses. Up until now we have constructed $\hypset$ using ancestral samples, following \citet{eikema-aziz-2020-map}. Strategies like nucleus sampling and beam search are known empirically to produce higher quality translations than ancestral sampling and might therefore also enumerate outcomes that have high expected utility. %
We explore ancestral sampling, nucleus sampling and beam search. In a hyperparameter search we found $p=0.7$ for nucleus sampling to work best. For beam search we use a length penalty of 1.2 (ne) or 0.6 (de, ro). We compare each strategy by the expected BEER values of the translations generated, using accurate estimates of expected BEER (using 1,000 samples for MC estimation). We show results in Figure~\ref{fig:alternative-hypothesis-spaces}.

We find ancestral sampling to produce hypotheses across the entire range of expected BEER scores. Nucleus sampling and beam search generally produce translations at the higher end of expected BEER. Therefore, these seem more suitable for generating effective $\hypset$ at smaller $N$. Nucleus sampling seems to lead to the largest proportion of high expected utility translations across language pairs. %
Beam search has a noticeably high proportion of poor translations for English-Nepali, a low-resource language pair where mode-seeking search has been observed to be less reliable. Results in the opposite direction were similar. We explore both nucleus sampling and beam search for constructing $\hypset$ in the next experiment, as well as combining all three strategies together.

\subsection{Coarse-to-Fine MBR}
\label{sec:experiments:coarse-to-fine-mbr}
We now turn to the coarse-to-fine procedure (\mbrcf) described in Section \ref{sec:method}. 

\subsubsection{Choice of Proxy Utility}
\label{sec:proxy-utilities}
\begin{figure*}
    \centering
    \includegraphics[width=\textwidth]{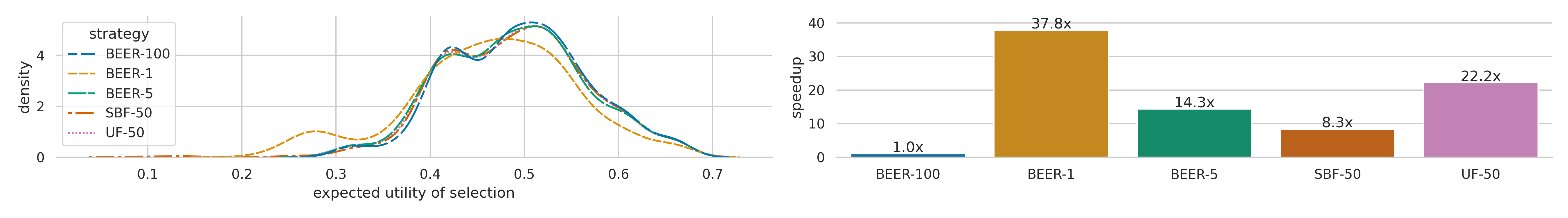}
    
    \caption{Comparison of proxy utilities on English to German: BEER using 1, 5 or 100 samples for MC estimation, and  unigram F1 (UF) and skip-bigram F1 (SBF) each using 50 samples for MC estimation. We use each proxy utility to filter a top-20 from 100 ancestral samples. We show the resulting expected target utilties (BEER, an accurate estimate) (left), as well as a runtime comparison (right). Results are aggregated over 100 source sequences.}
    \label{fig:proxy-utilities}
\end{figure*}

We compare various proxy utilities by their effectiveness as filtering strategies in obtaining high expected utility sets, where we again use accurate estimates of expected utility using 1,000 samples for MC estimation. We filter the top-20 hypotheses from an initial 100 hypotheses obtained using ancestral sampling. This ensures a high variety of expected utilities in the initial set. We also compare each proxy utility on their runtime performance. We compare both cheap estimates of expected BEER using either 1 or 5 samples for MC estimation (BEER-1 and BEER-5 respectively) as well as cheap-to-compute proxy metrics: unigram F1 using 50 samples for MC estimation (UF-50) and skip-bigram F1 using 50 samples for MC estimation (SBF-50).\footnote{Skip-bigrams are bigrams that do not enforce adjacency.}
We use expected BEER using 100 samples for MC estimation (BEER-100) as a reference point. See our results on the English-German system in Figure~\ref{fig:mbr-filter-strategies}.

We surprisingly find nearly all strategies to lead to equally good filtered sets as BEER-100 in terms of expected BEER of the filtered set.
The only strategy that performs slightly worse than the others is BEER-1, which is likely too noisy to be a reliable filtering strategy.
We observed very similar results for the other five language pairs. 
In terms of runtime performance we find BEER-1 to be fastest followed by UF-50 at a 22.2x performance increase over BEER-100.\footnote{Our Python implementations of unigram and skip-bigram F1 are not optimized and we deem it likely that a greater speed-up is possible with a more efficient implementation.}
In follow-up experiments, we will use UF-50 as a proxy utility, providing high quality filtered sets at good runtime performance.

\subsubsection{Coarse-to-Fine MBR Results}
\definecolor{Gray}{gray}{0.95}
\newcolumntype{a}{>{\columncolor{Gray}}r}
\begin{table}[t]
    \centering
    \footnotesize
    \setlength{\tabcolsep}{0.5em}
    \begin{tabular}{ll ar ar ar}
    \toprule
    & &  
     \multicolumn{2}{c}{en-de} &  \multicolumn{2}{c}{en-ro} &  \multicolumn{2}{c}{en-ne}  \\ \cmidrule(r{1pt}){3-4}\cmidrule(r{1pt}){5-6}\cmidrule{7-8}
    MBR & $\bar{\mathcal H}$ & {\scriptsize BEER} & {\scriptsize BLEU} &  {\scriptsize BEER}  & {\scriptsize BLEU}  &  {\scriptsize BEER}  & {\scriptsize BLEU}  \\  \midrule
    NxS & N & 64.3 & 38.0 & 54.9 & 21.4 & 38.9 & 3.6\\
    C2F & N & +1.1 & +1.9 & +0.4 & +0.2 & +0.4 & +0.2\\
     & B & +0.9 & +1.5 & \bf{+0.5} & \bf{+0.5} & +0.5 & \bf{+0.5}\\
     & all & \bf{+1.3} & +2.4 & \bf{+0.5} & +0.4 & \bf{+0.6} & \bf{+0.5}\\
    \midrule
    BS & - & +0.9 & \bf{+2.8} & -0.1 & +0.1 & -0.8 & +0.2\\
    \midrule\midrule
    & &  
     \multicolumn{2}{c}{de-en} &  \multicolumn{2}{c}{ro-en} &  \multicolumn{2}{c}{ne-en}  \\ \cmidrule(r{1pt}){3-4}\cmidrule(r{1pt}){5-6}\cmidrule{7-8}
    MBR & $\bar{\mathcal H}$  &  {\scriptsize BEER}  & {\scriptsize BLEU}  &  {\scriptsize BEER}  & {\scriptsize BLEU}  &  {\scriptsize BEER}  & {\scriptsize BLEU}  \\  \midrule
    NxS & N & 64.8 & 38.7 & 58.5 & 28.0 & 43.1 & 6.3\\
    C2F & N & +0.9 & +1.1 & +0.5 & +0.7 & +0.5 & +0.2\\
     & B & \bf{+1.0} & \bf{+1.5} & \bf{+0.7} & \bf{+1.2} & +0.5 & \bf{+0.9}\\
     & all & \bf{+1.0} & +1.4 & +0.6 & +1.1 & \bf{+0.8} & +0.8\\
    \midrule
    BS & - & +0.5 & +1.2 & -0.0 & +0.8 & -1.0 & +0.4\\    \bottomrule
    \end{tabular}
    \caption{Comparing \mbrns, \mbrcf and beam search (BS) in terms of BEER and BLEU performance. We use BEER as utility, UF-50 as proxy utility, set top-$T=50$ and use $L=100$ samples for MC estimation. We use various strategies for constructing $\hypset$: 405 nucleus samples (N), the 405-best list from beam search (B) and combining both of these along with 1,005 ancestral samples (all). We use $S=13$ in \mbrns to mimic the computational cost of \mbrcf at $N=405$. The last row shows standard beam search performance using a typical beam size of 4 or 5 depending on the language. MBR results are averaged over 3 runs. Standard deviations for BEER/BLEU scores are below 0.1/0.2 (NxS), 0.1/0.1 (C2F) and 0 (BS).}
    \label{tab:comparison-c2f-beam}
\end{table}

In Table~\ref{tab:comparison-c2f-beam} we compare \mbrcf with \mbrns using $N=405$ nucleus samples ($p=0.7$) to construct the hypothesis space. We filter the top-$T=50$ hypotheses using UF-50 as proxy utility and use $L=100$ samples for MC estimation of the top-set, following our findings in Sections~\ref{sec:proxy-utilities} and~\ref{sec:experiments:n-by-s-mbr} respectively. For \mbrns we set $S=13$ to roughly match the amount of computation available to \mbrcf, based on a 22.2x speed-up of UF-50 relative to BEER-100 observed in Figure~\ref{fig:proxy-utilities}. We find that across language pairs \mbrcf consistently outperforms \mbrns showing improvements between +$0.4$ and +$1.1$ BEER and +$0.2$ to +$1.9$ BLEU. \mbrcf thus is effective at obtaining higher translation quality than \mbrns at the same amount of computation available for MBR.

We also explore the effects on translation quality of changing and combining strategies for constructing $\hypset$. We find that using a beam of $N=405$ (using the same length penalty as in Section~\ref{sec:experiments:alternative-hypothesis-spaces}) to construct $\hypset$ produces better results than nucleus sampling for most language pairs. Notably, re-ordering a large beam considerably improves over standard beam search decoding (using the usual beam size of 5 (ro, ne) or 4 (de)) for all language pairs in terms of BEER and for most language pairs in terms of BLEU scores. Combining all strategies for creating hypothesis spaces: ancestral sampling, nucleus sampling and beam search leads to the best results overall. For all language pairs both BEER and BLEU scores either improve or remain similar. This is more empricial evidence that expected utility is a robust and reliable criterion for picking translations: enlarging the hypothesis space or improving MC estimation under reasonable choices of hyperparameters seemingly never unreasonably hurts translation quality, but generally improves it. 

\paragraph{A Multi-Reference Test Set} We also test three systems from Table~\ref{tab:comparison-c2f-beam} (NxS, C2F and beam search) on a multi-reference test set. We use the English to German systems trained on WMT18 news data and translate \texttt{newstest2021}, which has three separate translations for each source sentence (we use translators A, C and D). We show results in Table~\ref{tab:multi-ref}. We find a similar pattern to that of Table~\ref{tab:comparison-c2f-beam}. \mbrcf greatly outperforms \mbrns given the same amount of available compute (see Section~\ref{sec:experiments:coarse-to-fine-mbr}) for details). \mbrcf outperforms beam search results in terms of BEER, but is much closer to beam search this time in terms of BLEU.

\subsection{Runtime}
We measure runtime performance on hypothesis generation, sampling for MC estimation of expected utilities and decoding time seperately for various algorithms explored in this work on the English to German language pair. We run all experiments on an Intel Xeon Bronze 3104 Processor and a single NVIDIA GeForce 1080Ti GPU. For generating samples and beam search outputs we set the batch size to as large as possible, constrained by the available GPU memory. MBR using BEER as utility runs on CPU, while sampling and beam search run on GPU. We mimic the \mbrnn and \mbrcf setups from Table~\ref{tab:comparison-c2f-beam} using a hypothesis space of 405 nucleus samples. We also additionally include runtime results for \mbrnn with $N=405$ and a more expensive \mbrns variant with $S=100$ (NxS$_\text{large}$). For beam search we report results for a beam size of 4, as has been used throughout the paper for this language pair. Results are shown in Table~\ref{tab:runtime}. As can be seen, collecting hypotheses and unbiased sampling makes up for a large part of the total decoding time in MBR algorithms. We do note that sampling operations are easily parallelisable and can be split across multiple GPUs when available. In terms of the decoding time itself, we can see that we greatly reduced the amount of computation needed to perform MBR going from 23,156 seconds of decoding time for \mbrnn to only 726 seconds of decoding time for \mbrcf. This can be attributed to the great reduction in number of utility calls in our proposed approximations.

\begin{table}[]
    \centering
    \footnotesize
    \setlength{\tabcolsep}{0.5em}
    \begin{tabular}{l ll}
    \toprule
    \texttt{newstest2021}  & BEER & BLEU \\
    \midrule
    NxS  & 63.4 & 40.9 \\
    C2F  & \bf{64.5} & 42.8 \\
    \midrule
    BS & 63.7 & \bf{43.0} \\
    \bottomrule
    \end{tabular}
    \caption{English to German \mbrns and \mbrcf results on the \texttt{newstest2021} multi-reference test set. We use $N=405$ nucleus samples as hypothesis space and use the same hyperparameters as in Table~\ref{tab:comparison-c2f-beam}.}%
    \label{tab:multi-ref}
\end{table}

\begin{table}[]
    \centering
    \footnotesize
    \setlength{\tabcolsep}{0.5em}
    \begin{tabular}{l lllll}
    \toprule
    MBR & hyp. generation & sampling & decoding\\
    \midrule
    NxN & 6,241s & 7,739s & 23,156s\\
    NxS & 6,241s & 383s & 746s\\
    NxS$_\text{large}$ & 6,241s & 1,825s & 5,358s\\
    C2F & 6,241s & 1,825s  & 726s\\
    \midrule
    BS & -  & - & 194s\\
    \bottomrule
    \end{tabular}
    \caption{A runtime comparison of MBR variants and beam search. We separate the time taken for \one hypothesis generation \two sampling (for estimation of expected utility) and \three running the decoder itself. We use $N=405$ nucleus samples, $S=13$ and S$_\text{large}=100$ ancestral samples for NxS variants, and the hyperparameter settings for C2F as used in Table~\ref{tab:comparison-c2f-beam}.}
    \label{tab:runtime}
\end{table}

\section{Related Work}

In recent NMT literature MBR has started being explored either in combination with MAP decoding or replacing it altogether. \citet{stahlberg-etal-2017-neural} adapt lattice minimum Bayes risk decoding~\citep{tromble-etal-2008-lattice} on SMT translation lattices to be incorporated in left-to-right beam search decoding in NMT, thereby proposing a hybrid decoding scheme. They adapt lattice MBR to work on partial hypotheses and perform beam search to find translations that are both high probability under the NMT model and have high expected utility under the SMT model.
\citet{shu-nakayama-2017-later} also combine beam search with MBR decoding to find low risk hypotheses, after which they re-rank all hypotheses with MBR again. They report having to restrict the number of hypotheses as not to degrade the effectiveness of MBR re-ranking, a finding that is likely due to biased estimation of expected utility, as in our work we find that increasing the number of hypotheses always improves translation quality.
\citet{blain2017exploring} explore the quality of \nbest lists obtained from beam search in NMT models and find that while MAP  is not a good criterion for ranking the resulting hypotheses, re-ranking using MBR with BEER as a utility leads to improvements on top of standard beam search decoding (with a small beam size), in terms of both BLEU scores as well as human evaluation scores.
\citet{borgeaud-emerson-2020-leveraging} approach decoding from a voting theory perspective and derive a decoding strategy similar to MBR. They explore a range of utility functions, achieving similar BLEU scores to beam search, but showing improvements in terms of length, diversity and human judgement.

All of the above works make use of beam search to provide both the hypothesis space as well as to make a biased estimate of expected utility.
\citet{eikema-aziz-2020-map} are the first work in NMT that propose to use sampling from the model to both make unbiased estimates of expected utility, the importance of which we confirm in experiments, and to form the hypothesis space. The authors only explore \mbrnn, however, and never explore hypothesis spaces larger than $N=30$ samples. We show that it is beneficial to scale MBR to much larger hypothesis spaces and that it can be beneficial to construct them using mode-seeking strategies.
\citet{muller2021understanding} study the properties of the sampling-based algorithm proposed in \citet{eikema-aziz-2020-map} and explore hypothesis spaces up to a size of $N=100$ as well as multiple utility functions. They find that MBR decoding outputs exhibit a similar but smaller bias towards short translations and frequent tokens compared to beam search, but do observe that this is dependent on the choice of utility function. They further find that MBR decoding mitigates spurious copying and hallucinations under domain shift. Similar to our work, they find that MBR decoding scales well with larger hypothesis spaces and better estimation of expected utility.
\citet{freitag_mbr} explore the use of large hypothesis spaces and a range of utilities, including neural utilities, on the \mbrnn approximation. They find that using BLEURT as utility leads to significantly better translations in a human evaluation, while producing considerably lower probability translations.

We provide a more extensive overview of historical approximations to the MBR objective as well as an overview of alternatives for tackling the inadequacy of the mode in Appendix~\ref{app:additional-related-work}.

\section{Conclusion}

We have shown MBR to be a robust decision rule for NMT that can find high quality translations. In particular, we have found that MBR, under reasonable hyperparameter choices, generally leads to improved translation quality with more computation (\ie, searching a larger search space and/or using more samples for more accurate MC estimation). Big challenges in decoding with MBR are constructing the hypothesis space and keeping computational cost of estimating expected utility tractable. We have proposed effective strategies for both, by exploring more efficient ways of forming the hypothesis space and proposing an approximation to MBR that is linear in the size of this hypothesis space. 
Our coarse-to-fine MBR procedure is able to considerably reduce the number of calls to the utility function without compromising translation quality.
We have shown that sampling-based MBR in general can outperform beam search on all the language pairs we explored and can continue to improve with better and more accurate search. 
We believe sampling-based MBR to be a promising, albeit still more expensive, alternative to beam search decoding. 
Unlike beam search, where it is not obvious how to further improve translation quality, sampling-based MBR is likely to benefit from improvements of different aspects of the algorithm. 
We believe fruitful avenues of research to be among \one clever algorithms for constructing hypothesis spaces, \two more robust estimates of expected utility using fewer samples, \three use of modern neural utilities and \four improving the modelling capacity of NMT systems.
We hope that this work motivates researchers and practitioners to make more conscious considerations of the choice of decision rule and that it paves the way for use of tractable sampling-based MBR decoding in NMT.

\section*{Limitations}
This work has proposed a number of algorithms for more efficient decoding under the minimum Bayes risk decision rule. However, in terms of runtime performance MBR decoding is still outperformed by beam search. MBR will likely always be more expensive than current applications of beam search, in which very small beam sizes are employed, since on top of generating translation candidates, MBR decoding will potentially need a separate set of samples for estimating expected utility, and perform additional computations in the form of utility assessments. While this currently makes MBR less attractive in real-time translation scenarios, we believe that the demonstrated scalability and robustness of the decoding objective makes MBR interesting in scenarios in which translation speed is not the highest priority. Furthermore, continued research into algorithmic improvements to MBR approximations and optimized implementations of existing algorithms may make MBR attractive in real-time translation in the future.

MBR also relies on a utility function, a hyperparameter to the decision rule (decoding algorithm). On the one hand, this allows us to inject some domain expertise into the decoding algorithm. On the other hand, in machine translation, we do not have a gold-standard metric that we trust to judge translation quality perfectly. This means we will have to choose a utility that we know is suboptimal, and may have peculiarities such as bad hypotheses that exploit certain aspects of the utility to be ranked unreasonably high. Nonetheless, it is unlikely that the NMT model puts a lot of mass on such translations, reducing the likelihood of encountering such situations. We believe there are also positives to incorporating a utility function into the decoding algorithm: MBR can benefit from advances in the field of machine translation evaluation, as some recent works have already exploited \citep{freitag_mbr,fernandes-etal-2022-quality}.

Finally, current MBR algorithms do not permit incremental generation of translations. A translation hypothesis can only be assessed once it’s fully generated by the NMT model. This is a bottleneck to its speed and doesn’t make optimal use of the factorisation of modern-day NMT systems. We do think this is a promising direction for future work.

\iffinal
\section*{Acknowledgements}

\begin{wrapfigure}[2]{l}{0.12\linewidth}
\vspace{-13pt}
\includegraphics[width=0.08\textwidth]{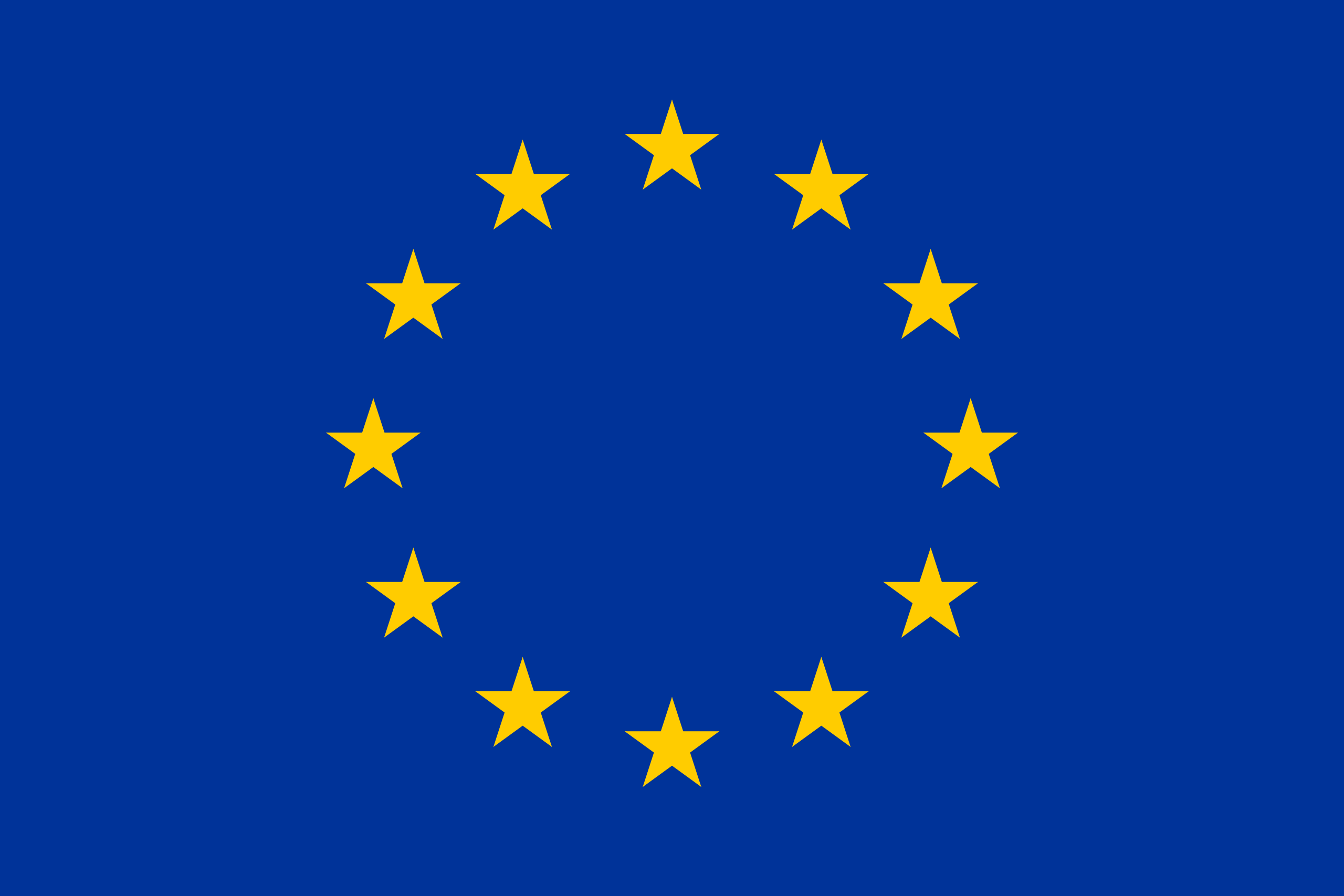}
\end{wrapfigure}

This project has received funding from the European Union's Horizon 2020 research and innovation programme under grant agreement No 825299 (GoURMET).  
\fi

\bibliography{anthology,custom,coling}

\appendix
\section{Additional Related Work}
\label{app:additional-related-work}

\subsection{Approximations to MBR}

Most instances of MBR decoding in machine translation, from the original work of \citet{kumar-byrne-2004-minimum} to recent instances in NMT \citep{stahlberg-etal-2017-neural,shu-nakayama-2017-later,blain2017exploring}, approximate the objective function by computing expectations not \wrt the model distribution, but rather, \wrt a proxy distribution. This proxy is obtained by enumeration via beam-search of a subset of the sample space (\eg, a \nbest list), and renormalisation of the probabilities of the outcomes in this subset. 
This has the undesirable effect of exaggerating differences in probability due to underestimation of the normalisation constant, and, like MAP decoding, it over-represents pathologies around the mode.
Similarly, most prior work uses mode-seeking search to explore a tractable subset of the hypothesis space. 
Mode-seeking approximations bias the decoder towards the mode making MBR decoding less robust to idiosyncratic outcomes in the hypothesis space \citep{eikema-aziz-2020-map}.%
This is in stark contrast with our work, where we sample from the model to construct unbiased estimates of expected utility, as well as to enumerate a tractable hypothesis space.

There are cases in statistical machine translation (SMT) where the computation of expected utility can be factorised along a tractable directed acyclic graph (DAG) via dynamic programming \citep{tromble-etal-2008-lattice,zhang-gildea-2008-efficient,denero-etal-2009-fast,kumar-etal-2009-efficient}. In such cases, the DAG contains a much larger subset of the sample space than any practical \nbest list, still some  pruning is necessary to construct a compact DAG containing only the most probable outcomes. These strategies are only available for models and utility functions that make strong Markov assumptions. For example, \citet{tromble-etal-2008-lattice} and \citet{denero-etal-2009-fast} develop linearisation strategies for BLEU, and \citet{zhang-gildea-2008-efficient} maximise expected trigram counts as a proxy to BLEU proper. The idea of utilising a proxy utility is something we also explore in this paper, though only as an intermediate step to decoding with the target utility.

In some (rarer) cases, unbiased (or asymptotically unbiased) samples have been used to approximate the MBR objective and/or to reduce the search space. For example, \citet{stanojevic-simaan-2015-reordering} use ancestral sampling in MBR decoding for permutation-trees-based reordering models, and \citet{arun-etal-2009-monte} use Gibbs sampling for MBR decoding in phrase-based MT.
Unbiased samples for estimation of expected utility or exploration of a tractable hypothesis space are simply not common in machine translation. In SMT, the reason is a technical one, most SMT models are not based on a left-to-right factorisation of the joint distribution, thus unbiased sampling requires MCMC \citep{denero-etal-2008-sampling,blunsom-etal-2009-gibbs} or expensive adaptive rejection sampling  \citep{aziz-etal-2013-investigations}. 
This limitation does not extend to NMT models, but NMT most likely simply inherited from SMT the practice of using beam-search-based approximations, at least until the work of \citet{eikema-aziz-2020-map}.

\subsection{Tackling the Inadequacy of the Mode}

\citet{eikema-aziz-2020-map} link the inadequacy of the mode in NMT to the entropy of the conditional distribution, or, more precisely, to the fact that NMT models tend to spread probability mass over large subsets of the sample space \citep{ott-etal-2018-analyzing}. 
It is plausible that strategies to concentrate probability mass (\eg, reducing entropy or pruning the support of the model) will do so by making inadequate translations less probable. 
For example, \citet{forster-etal-2021-searching} find that the inadequacy of the mode problem does not seem to affect sequence-to-sequence models of morphological inflection, an essentially deterministic task, whose combinatorial space is built upon a smaller vocabulary (\ie, characters instead of sub-word units), and whose observations are typically very short (\ie, words rather than sentences). 
\citet{peters-martins-2021-smoothing} train sparse sequence-to-sequence models \citep{peters-etal-2019-sparse} which assign zero probability to many outcomes dramatically reducing the support of the conditional distribution over complete sequences. They show that sparsity leads to inadequate candidates such as the empty string being pruned out of the support.
They also find that label smoothing increases the rate at which the empty string is more probable than the beam-search output.

\citet{meister-etal-2020-beam}  interprets the algorithmic approximations of beam search as an inductive bias towards outputs with uniform information density~\citep{UID}. They develop variants of beam search where this preference is a tunable hyperparameter and show that deviating from the mode with this type of bias can lead to improved translation quality. 
Another way to deviate from the mode is to augment the decoding objective with an auxiliary model. 
\citet{li2016mutual} re-rank a \nbest list using a combination of two model probabilities, namely, $p_{Y|X}(h|x, \theta_{\text{fwd}})$ and $p_{X|Y}(x|h, \theta_{\text{bwd}})$. They think of this as maximising the mutual information (MI) between source and translation. The motivation is that the target-to-source component will push against inadequate candidates, as those are unlikely to be mapped back to the source with high probability.
\citet{bhattacharyya-etal-2021-energy} find that 100 samples from an NMT model contain better candidates (measured in terms of BLEU) than the output of beam search (an observation \citet{eikema-aziz-2020-map} also make based on 30 samples and METEOR, instead). They propose to rerank these samples using an energy-based model trained to order candidates as sentence-BLEU would. 
Like these works, sampling-based MBR decoding, can be seen as a form of \emph{explore and rank} approach, however, the ranking function in MBR is derived from the NMT model itself, whereas both MI- and EBM-based re-ranking involve an auxiliary trained model. For the EBM, in particular, in the limit of a too large hypothesis space, the beliefs of the NMT model are completely overwritten by the EBM. MBR, instead, does not overwrite the model's beliefs, it re-expresses those beliefs in terms of utility.

\citet{leblond2021machine} recast NMT as a reinforcement learning problem and learn both a policy (\ie, a mechanism to explore the space of translations one word at a time from left-to-right) and a value function (\ie, an estimate at the expected reward of finishing a given prefix translation). For reward they investigate what they call privileged metrics, which require access to references (\eg, sentence-level BLEU), and unprivileged metrics, which do not use references but access the source (\eg, a quality estimation score).
Compared to sampling-based MBR, their work tightly integrates search and value estimation, thus going beyond ranking a fixed set of candidates.  The objective function of MBR can be thought of as an `unprivileged metric' in their terminology, one that is based on the NMT model itself (and a choice of utility).
But, the policy in sampling-based MBR (\ie, the NMT model) is not trained to be aware of the evaluation metric.

\section{Comparing Target Utilities}
\label{app:additional-results}
\begin{table}[]
    \centering
    \scriptsize
    \pgfplotstableread{data/utilities.tex}\loadedtable
\pgfplotstablesort[
    sort key={order}, 
    sort cmp={string <},
]{\sortedtable}{\loadedtable}
\pgfplotstabletypeset[
    columns={%
        task,
        utility,
        beer,
        bleu,
        meteor,
        chrfpp
    },
    columns/task/.style={
        column name={Task},
        string type,
        column type=l,
    },
    columns/utility/.style={
        column name={Utility},
        string type,
        column type=l,
        string replace={beer}{BEER},
        string replace={bleu}{sentence-BLEU},
        string replace={meteor}{METEOR},
        string replace={chrf++}{ChrF++},
    },
    columns/meteor/.style={
        column name={METEOR},
        column type=r,
        precision=1,
        clear infinite,
        fixed,
        fixed zerofill,
    },
    columns/beer/.style={
        column name={BEER},
        column type=r,
        precision=1,
        clear infinite,
        fixed,
        fixed zerofill,
    },
    columns/bleu/.style={
        column name={BLEU},
        column type=r,
        precision=1,
        clear infinite,
        fixed,
        fixed zerofill,
    },
    columns/chrfpp/.style={
        column name={ChrF++},
        column type=r,
        precision=1,
        clear infinite,
        fixed,
        fixed zerofill,
    },
    columns/bleurt/.style={
        column name={BLEURT},
        column type=r,
        precision=1,
        clear infinite,
        fixed,
        fixed zerofill,
    },
    every last row/.style={
        after row={
            \bottomrule
        }
    },
    every head row/.style={ 
        before row={
            \toprule 
        },
        after row=\midrule,
    },
    every row no 4/.style={before row=\midrule},
    every row no 8/.style={before row=\midrule},
    every row no 12/.style={before row=\midrule},
    every row no 16/.style={before row=\midrule},
    every row no 20/.style={before row=\midrule},
    every row 0 column 2/.style={
        postproc cell content/.style={
          @cell content/.add={$\bf}{$}
        }
    },
    every row 1 column 3/.style={
        postproc cell content/.style={
          @cell content/.add={$\bf}{$}
        }
    },
    every row 2 column 4/.style={
        postproc cell content/.style={
          @cell content/.add={$\bf}{$}
        }
    },
    every row 3 column 5/.style={
        postproc cell content/.style={
          @cell content/.add={$\bf}{$}
        }
    },
    every row 4 column 2/.style={
        postproc cell content/.style={
          @cell content/.add={$\bf}{$}
        }
    },
    every row 5 column 3/.style={
        postproc cell content/.style={
          @cell content/.add={$\bf}{$}
        }
    },
    every row 6 column 4/.style={
        postproc cell content/.style={
          @cell content/.add={$\bf}{$}
        }
    },
    every row 7 column 5/.style={
        postproc cell content/.style={
          @cell content/.add={$\bf}{$}
        }
    },
    every row 8 column 2/.style={
        postproc cell content/.style={
          @cell content/.add={$\bf}{$}
        }
    },
    every row 9 column 3/.style={
        postproc cell content/.style={
          @cell content/.add={$\bf}{$}
        }
    },
    every row 10 column 4/.style={
        postproc cell content/.style={
          @cell content/.add={$\bf}{$}
        }
    },
    every row 11 column 5/.style={
        postproc cell content/.style={
          @cell content/.add={$\bf}{$}
        }
    },
    every row 12 column 2/.style={
        postproc cell content/.style={
          @cell content/.add={$\bf}{$}
        }
    },
    every row 13 column 3/.style={
        postproc cell content/.style={
          @cell content/.add={$\bf}{$}
        }
    },
    every row 14 column 4/.style={
        postproc cell content/.style={
          @cell content/.add={$\bf}{$}
        }
    },
    every row 15 column 5/.style={
        postproc cell content/.style={
          @cell content/.add={$\bf}{$}
        }
    },
    every row 16 column 2/.style={
        postproc cell content/.style={
          @cell content/.add={$\bf}{$}
        }
    },
    every row 16 column 3/.style={
        postproc cell content/.style={
          @cell content/.add={$\bf}{$}
        }
    },
    every row 18 column 3/.style={
        postproc cell content/.style={
          @cell content/.add={$\bf}{$}
        }
    },
    every row 18 column 4/.style={
        postproc cell content/.style={
          @cell content/.add={$\bf}{$}
        }
    },
    every row 19 column 5/.style={
        postproc cell content/.style={
          @cell content/.add={$\bf}{$}
        }
    },
    every row 20 column 2/.style={
        postproc cell content/.style={
          @cell content/.add={$\bf}{$}
        }
    },
    every row 20 column 3/.style={
        postproc cell content/.style={
          @cell content/.add={$\bf}{$}
        }
    },
    every row 22 column 4/.style={
        postproc cell content/.style={
          @cell content/.add={$\bf}{$}
        }
    },
    every row 23 column 5/.style={
        postproc cell content/.style={
          @cell content/.add={$\bf}{$}
        }
    },
]\sortedtable
    \caption{Comparing BEER, sentence-BLEU, METEOR and ChrF++ as utility functions in \mbrns using $N=405$ and $S=100$.}
    \label{tab:app:utilities}
\end{table}

We compare a number of utility functions for use in MBR decoding. In principle any function that measures some notion of similarity across sequences and can be reliably assessed on the sentence-level is suitable as a utility function for MBR. As BLEU is the predominant automatic evaluation metric on which translation quality is assessed, we experiment with a smoothed version of BLEU~\citep{bleu} that can work on the sentence-level: sentence-BLEU~\citep{chen-cherry-2014-systematic} using the default parameters in \citet{post-2018-call}. We further try METEOR~\citep{meteor} as this was used in \citet{eikema-aziz-2020-map} and showed good results.\footnote{We use a slightly different version of METEOR than in \citet{eikema-aziz-2020-map}. We use language-specific versions rather than a language-agnostic version used in that work.} BEER~\citep{stanojevic-simaan-2014-fitting} is a character-based metric that has shown to correlate well with human judgements in many WMT metrics tasks~\citep{machacek-bojar-2014-results,stanojevic-etal-2015-results,bojar-etal-2016-results}. Finally, we also explore ChrF++~\citep{popovic-2017-chrf}, another character based metric that is an improved version of ChrF~\citep{popovic-2015-chrf}.

We perform \mbrns with $N=405$ and $S=100$ in order to perform the comparisons. We measure the performance of each utility on BEER, BLEU, METEOR and ChrF++. Our results are shown in Table~\ref{tab:app:utilities}. As expected, using a certain utility achieves the best performance under the lens of that metric as well. Sometimes we find a small deviation from this when BEER or METEOR outperforms sentence-BLEU in terms of BLEU score. This is likely due to sentence-BLEU only being an approximation to BLEU itself. We find that overall BEER seems to do best across metrics followed by ChrF++. One attempt to quantify this more clearly is by normalize the scores per language pair and evaluation metric compared to the maximum score obtained by the best scoring system for that metric and language pair. This leads to the following average performances per evaluation metric: BEER 0.978, METEOR 0.968, ChrF++ 0.964, and sentence-BLEU 0.955. This indeed shows a slight edge of BEER over the other utilities tested in pushing scores across our evaluation metrics. Herefore, in the main paper, we have used BEER as the utility of choice. The finding that BEER works well as a utility function in MBR was also made before in the work of \citet{blain2017exploring}.

\end{document}